\documentclass[runningheads]{llncs}
\usepackage{graphicx}
\usepackage{comment}
\usepackage{amsmath,amssymb}
\usepackage{color}

\usepackage{mathtools}
\usepackage{bm}
\usepackage{multirow}
\usepackage{array}
\usepackage[linesnumbered,ruled]{algorithm2e}
\usepackage{booktabs}
\usepackage{pifont} 

\usepackage[breaklinks=true,bookmarks=false]{hyperref}

\newcommand{\cmark}{\ding{51}}%
\newcommand{\xmark}{\ding{55}}%

\DeclareMathOperator*{\argmax}{arg\,max}

\DeclareMathOperator*{\softmax}{softmax}
\DeclareMathOperator*{\fc}{fc}

\newcommand{\model}{JVGN}
\newcommand{\modelspace}{JVGN }

\begin{document}
\pagestyle{headings}
\mainmatter
\def\ECCVSubNumber{4203}  

\title{Joint Visual Grounding \\ with Language Scene Graphs} 

\titlerunning{Joint Visual Grounding with Language Scene Graphs}
%
\author{Daqing Liu\inst{1} \and
Hanwang Zhang\inst{2} \and
Zheng-Jun Zha\inst{1} \and \\
Meng Wang\inst{3} \and
Qianru Sun\inst{4}}
\authorrunning{D. Liu, H. Zhang, Z.J. Zha, M. Wang, and Q. Sun}
%
\institute{University of Science and Technology of China \and
Nanyang Technological University \and
Hefei University of Technology \and
Singapore Management University \\
\email{liudq@mail.ustc.edu.cn, hanwangzhang@ntu.edu.sg, zhazj@ustc.edu.cn}\\ \email{wangmeng@hfut.edu.cn, qianrusun@smu.edu.sg}
}
\maketitle

\begin{abstract}
Visual grounding is a task to ground referring expressions in images, \textit{e.g.}, localize ``the white truck in front of the yellow one''. To resolve this task fundamentally, the model should first find out the contextual objects (\textit{e.g.}, the ``yellow'' truck) and then exploit them to disambiguate the referent from other similar objects by using the attributes and relationships (\textit{e.g.}, ``white'', ``yellow'', ``in front of'').
However, due to the lack of annotations on contextual objects and their relationships, existing methods degenerate the above joint grounding process into a holistic association between the expression and regions, thus suffering from unsatisfactory performance and limited interpretability.
In this paper, we alleviate the missing-annotation problem and enable the joint reasoning by leveraging the \textit{language scene graph} which covers both labeled referent and unlabeled contexts (other objects, attributes, and relationships).
Specifically, the language scene graph is a graphical representation where the nodes are objects with attributes and the edges are relationships. We construct a factor graph based on it and then perform marginalization over the graph, such that we can ground both referent and contexts on corresponding image regions to achieve the joint visual grounding (JVG).
Experimental results demonstrate that the proposed approach is effective and interpretable, \textit{e.g.}, on three benchmarks, it outperforms the state-of-the-art methods while offers a complete grounding of all the objects mentioned in the referring expression.

\end{abstract}

\section{Introduction}

\begin{figure}[t]
	\centering
	\includegraphics[width=\linewidth]{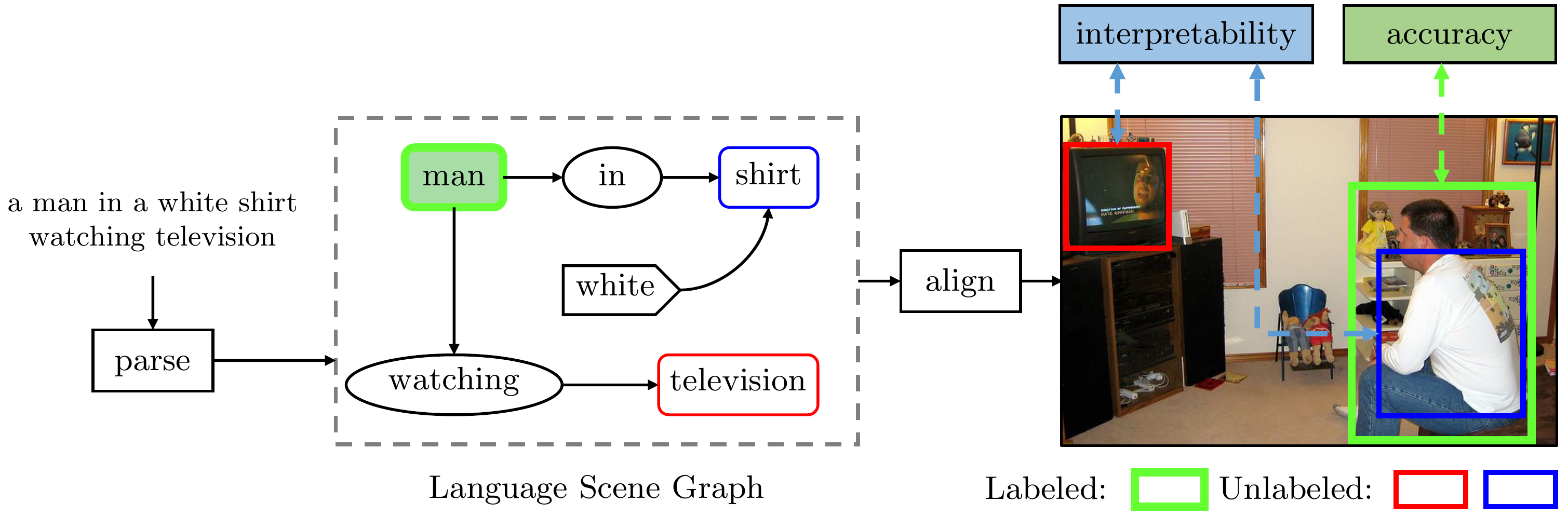}
	\caption{
		An intuitive illustration of the proposed JVGN approach. Language scene graph legends are: green shaded rectangle: referent node, colored rectangle: object node, arrow rectangle: attribute, oval: edge relationship. The same color of the bounding box and the node denotes a grounding. Thanks to the marginalization strategy on scene graphs, our approach can yield an efficient joint visual grounding accuracy as well as the context-based interpretability.
	}
	\label{fig:tisser}
\end{figure}

Grounding referring expressions in visual scenes is perhaps the most natural human control for AI, \textit{e.g.}, ``park the car beside the red sedan in front of the blue gate'' for a self-driving car~\cite{chen2018touchdown}, and ``who is the man in blue with a dress watch'' for a visual question answering (VQA) agent~\cite{antol2015vqa}. 
Beyond object detection~\cite{ren2015faster}, visual grounding task fundamentally requires the full understanding of the language compositions conditioned on the images (\textit{e.g.}, the linguistic meaning of ``beside'' and ``with'' connecting objects), and then use them as the guidance to distinguish the referent out of the contexts, especially those of the same class (\textit{e.g.}, ``the man'' vs. ``other men'').
Towards a better comprehension of the language compositions, it is crucial to identify each noun phrase as well as capture the relationships between them in an image.

However, due to the lack of annotations of contextual objects mentioned in the expression, most previous methods implicitly encode the context information into a latent representation by decomposing the language into a sequence or a tree, and only optimize the referent grounding results but not the global grounding for all objects.
Unfortunately, realizing this idea is very challenging due to the prohibitive cost of annotating a complete grounding for all possible expressions, as the number of multi-object articulation in the visual world is exponentially large.
It is also a common challenge for many other visual reasoning tasks such as VQA~\cite{johnson2017clevr}, image captioning~\cite{vinyals2017show}, and visual dialog~\cite{das2017visual}.
Therefore, given only the referent's ground-truth, almost all popular visual grounding methods lower the requirement of joint grounding and reasoning to a holistic association between the sentence and region features~\cite{hu2017modeling,zhang2018grounding,yu2018mattnet}.
For example, the state-of-the-art method~\cite{yu2018mattnet} coarsely models the triplet score of (subject, predicate, object), regardless of the sentence complexity, \textit{e.g.}, the ``object'' may still have its own sub-ordinate triplet decomposition and so on. As these methods violate the nature of visual reasoning, even for a correct grounding result, its inference may not be faithful and interpretable to the language composition, and thus it is poorly generalized to unseen expressions.

In fact, we are not the first to re-think the downside of the holistic models. Inspired by the success of neural module networks in synthetic VQA datasets \cite{hu2017learning,shi2018explainable,Johnson_2017_ICCV}, where the visual reasoning is guided by the question parsing trees, researchers attempt to localize the objects along the expression parsing trees for visual grounding. However, due to the difficulty in training the highly moving neural modules with the massively missing annotations of the contexts, they are either significantly under-performed~\cite{cirik2018using} or easily degenerated to holistic scores with limited interpretability~\cite{hong2019learning,cao2018interpretable}.

In this paper, we present a novel visual grounding framework that is able to do joint visual grounding (JVG) on both labeled referent and unlabeled contexts, \textit{i.e.}, all the objects mentioned in the sentence.
To obtain the semantic composition of a sentence, we use an off-the-shelf scene graph parser~\cite{schuster2015generating} to parse the sentence into a language scene graph, where a node is an entity object modified by attributes, and an edge is a relationship between two nodes (cf. Figure~\ref{fig:tisser}). 
Such a scene graph offers a graphical inductive bias~\cite{battaglia2018relational} for the joint grounding and reasoning. As detailed in Section~\ref{sec:3}, we align the image with scene graph by leveraging the
Conditional Random Field (CRF), where the image regions can be considered as the observational label space of the scene graph.
In particular, we describe the CRF by a Factor Graph where variables are objects and the unary and binary factors are single and pairwise vision-language association scores, respectively.
To train the model, we face the main challenge of missing the ground-truth of the contextual nodes.
We address this by the marginalization strategy with message passing which inherently and effectively enables the joint label propagation between labeled and unlabeled nodes.
Specifically, we marginalize out the joint distribution of unlabeled nodes
(\textit{e.g.}, $P$(``shirt'', ``television'') in Figure~\ref{fig:tisser}) in order to obtain the marginal probability of referent node (\textit{e.g.}, $P$(``man'')).
As the ground-truth of this node is given, the proposed model can be trained with KL-Divergence loss.
Thanks to the chain rule, the gradient back-propagates over the whole scene graph.
We call our overall model as Joint Visual Grounding Network (JVGN).
On three visual grounding benchmarks~\cite{yu2016modeling,mao2016generation}, we conduct extensive analysis and comparisons with the state-of-the-arts. Thanks to the fact that the proposed JVGN is a well-posed probabilistic graphical model, we can make the best of the two worlds: it consistently outperforms the popular holistic networks of low interpretability, while retains the high interpretability of structural models.

To summarize, our contributions are three-fold:
(1) A novel Joint Visual Grounding Network (JVGN) with scene graphs that incorporate linguistic structural information and graphical inductive bias;
(2) A marginalization strategy implemented with message passing on factor graphs that tackles the missing-annotation problem of contextual nodes on the scene graphs;
(3) Extensive quantitative and qualitative experiments on three visual grounding benchmarks, where our approach achieves both the state-of-the-art performance and the high interpretability.

\section{Related Work}

\subsection{Visual Grounding}
Visual grounding (\textit{a.k.a.}, referring expression comprehension~\cite{mao2016generation}) is a task to localize a region in an image, where the region is described by a natural language expression.
Early methods~\cite{hu2016natural,mao2016generation,yu2016modeling,luo2017comprehension,yu2017joint} are designed for both referring expression comprehension and generation tasks. They used the CNN-LSTM encoder-decoder structure to localize the region which could generate the sentence with maximum posteriori probability.
After that, discriminative methods~\cite{hu2017modeling,yu2018mattnet,yu2018rethining,zhang2018grounding,liu2019learning,hong2019learning} are widely used to solve the comprehension task independently. They usually compute a similarity score between the sentence and region, which can be discriminatively trained with cross-entropy loss~\cite{wang2016learning} or max margin loss~\cite{hu2016natural}.
To the language end, \cite{hu2017modeling,yu2018rethining} represented the language as a holistic embedding with soft attention~\cite{deng2018visual,zhuang2018parallel} mechanism.
\cite{hu2017modeling,yu2018mattnet,peng2019grounding} coarsely decompose the language into pre-defined triplets, \textit{e.g.}, (subject, relation, object)~\cite{hu2017modeling} and (subject, location, relation)~\cite{peng2019grounding}. Most recently, \cite{cirik2018using,liu2019learning,hong2019learning} transform the language into a tree structure and then perform visual reasoning along the tree in a bottom-up fashion.

However, most of the above approaches only focus on the referent grounding score calculation but neglect the contextual objects mentioned in the language, which is crucial to disambiguate the referent from other contextual objects, especially the objects of the same category. Thus, they easily rely on shallow correlations introduced by unintended biases~\cite{cirik2018visual}, rather than visual reasoning over the whole vision-language scene. To this end, we explicitly perform joint visual grounding on both labeled referent object and unlabeled contextual objects on a scene graph~\cite{schuster2015generating} parsed from the language.

\subsection{Scene Graphs}
Scene graphs have been widely used in vision-language tasks recently. Most existing works~\cite{zellers2018neural} rely on \textit{visual} scene graph detected from images, where nodes are visual object categories and edges are visual relationships.
Those visual scene graphs are limited by pre-defined object and relation categories, thus hardly aligned with natural language.
Therefore, our work leveraged the \textit{language} scene graph~\cite{schuster2015generating} parsed from a natural language sentence, where nodes are noun phrases and edges are dependency relations.
Similar to visual scene graphs, the language counterpart serves as a reasoning inductive bias~\cite{battaglia2018relational} that regularizes the model training and inference, which has been shown useful in image generation~\cite{johnson2018image} and captioning~\cite{yang2019caption}.

There are several works~\cite{johnson2015image,plummer2017phrase,zhang2014robust,zhang2012attribute} related to the language scene graph.
For image retrieval, \cite{johnson2015image} used a CRF to ground the scene graph to images for calculating the similarity between the image and sentence. For phrase grounding, \cite{plummer2017phrase} extracted several noun phrases and dependency relations from language as cues, and then localized each noun phrases by considering those cues jointly.
However, those above methods heavily relied on prohibitive annotations of each node, leading to limitations to real-world applications.
To this end, our work used Factor Graph as a framework to marginalize the unlabeled contexts into the labeled referent, making the whole scene graph end-to-end trainable.

\section{Approach}\label{sec:3}

\begin{figure*}[t]
	\centering
	\includegraphics[width=\linewidth]{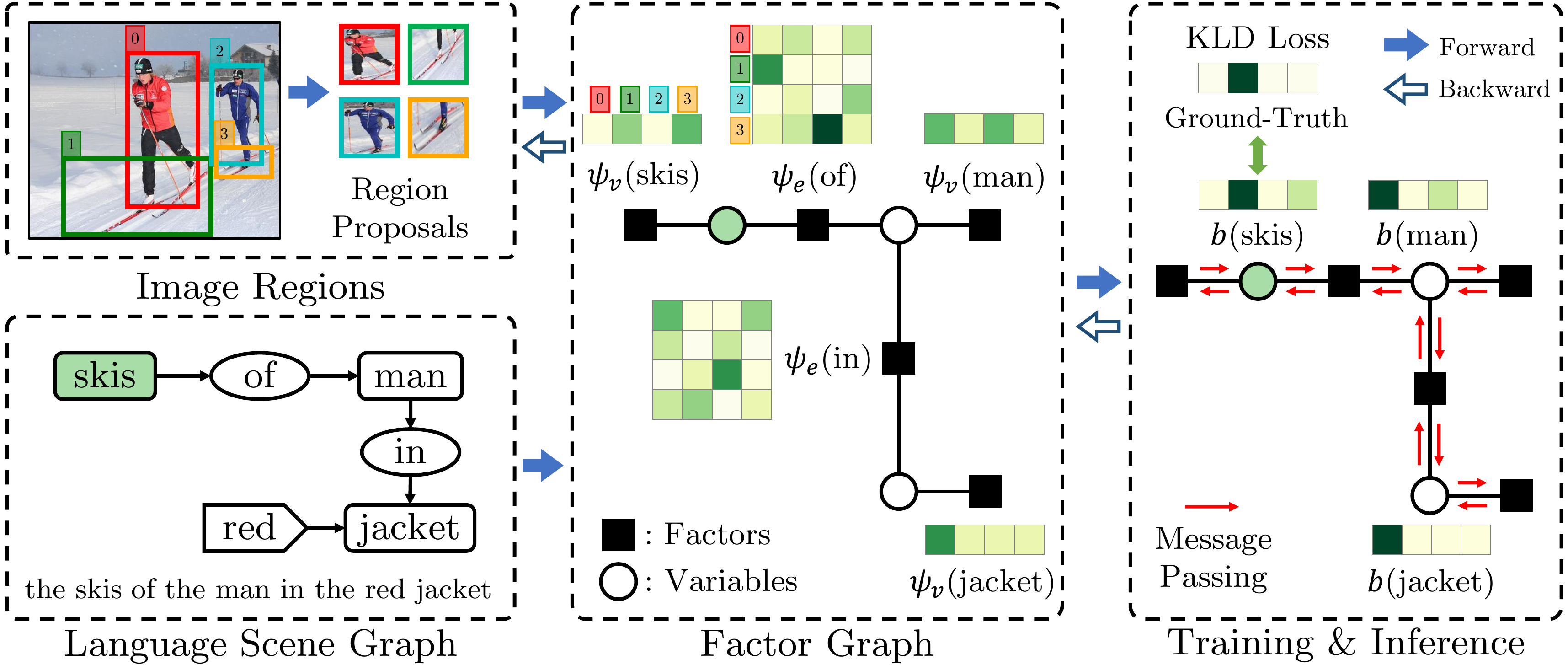}
	\caption{The overview of the proposed Joint Visual Grounding Network (\model). Given a natural language expression as input, we first parse it into a language scene graph. Then, we build a factor graph based on the scene graph and compute unary and binary potential functions for factors.
	Next, we perform the message passing over this graph to marginalize out contextual nodes.
	Finally, we train JVGN with the KLD loss between the ground-truth and the marginal probability on the referent node.
	}
	\label{fig:overview}
\end{figure*}

In this section, we first formulate the visual grounding problem in Section~\ref{sec:3.1}. Then, we introduce the proposed Joint Visual Grounding Network (\model) in Section~\ref{sec:3.2}. Finally, we detail how to train and infer the model in Section~\ref{sec:3.3}.

\subsection{Problem Formulation}\label{sec:3.1}
The visual grounding task aims to localize the referent region in an image given a referring expression.
For the given image $\mathcal{I}$, we represent it as a set of regions $\mathcal{B}=\{b_1, \cdots, b_N\}$, where $N$ is the number of regions. Fo the referring expression $\mathcal{L}$, we represent it as a sequence of words $\mathcal{L}=\{l_1, \cdots, l_L\}$, where $L$ is the length of sentence. The visual grounding can be formulated as a \textit{retrieval} problem that returns the most possible grounding from $\mathcal{L}$ to $\mathcal{B}$.
Here, we introduce a \textit{random variable} $\gamma: \mathcal{L} \rightarrow \mathcal{B}$, which denotes a grounding of the referent mentioned in $\mathcal{L}$ to a region in $\mathcal{B}$. Formally, we formulate the visual grounding problem as finding the optimal $\gamma^*$ as:
\begin{equation}
\label{eq:1}
    \gamma^* = \argmax_{\gamma} P(\gamma \mid \mathcal{L}, \mathcal{I}).
\end{equation}
In fact, almost all state-of-the-art models~\cite{hu2017modeling,yu2018mattnet,yu2018rethining,zhang2018grounding} can be formulated as Eq.~\eqref{eq:1} where the language composition is oversimplified, \textit{i.e.}, no joint grounding of all mentioned objects or compositional visual reasoning is taken into account. On the contrary, we believe that a principled visual grounding solution should be faithful to all the objects mentioned in the language.
Therefore, we reformulate the visual grounding task as an \textit{inference} problem over all objects, \textit{i.e.}, joint visual grounding.
Without loss of generality, we assume that there are $M$ groundings for $M$ objects mentioned in $\mathcal{L}$ and denote the first grounding $\gamma_1$ as the referent grounding.
Formally, searching for the optimal joint grounding can be formulated as:
\begin{equation}
\label{eq:2}
    \gamma_1^*,\gamma_2^*,...,\gamma_M^* = \argmax_{\gamma_1,...,\gamma_M} P(\gamma_1,\gamma_2,...,\gamma_M\mid\mathcal{L}, \mathcal{I}),
\end{equation}
where we can easily find out that the key is to model the joint probability of all the grounding $P(\gamma_1, ..., \gamma_M\mid\mathcal{L},\mathcal{I})$.

Next, we will detail the implementation of the joint probability using our proposed Joint Visual Grounding Network.

\subsection{Joint Visual Grounding Network}\label{sec:3.2}
The overview of the proposed Joint Visual Grounding Network (\model) is illustrated in Figure~\ref{fig:overview}.
First, we extract region features from the image, construct the language scene graph, and generate the language representations.
Second, we build a CRF model based on the scene graph and convert it into a Factor Graph with variables and factors.
Last, we perform the belief propagation over the factor graph to compute the marginal probability, which can be used to train our model and infer the joint grounding results for all the objects, including unlabeled contexts.

\subsubsection{Language Scene Graph}
The language scene graph, as shown in Figure~\ref{fig:overview}, is defined as $\mathcal{G} = (\mathcal{V}, \mathcal{E})$, where $\mathcal{V}=\{v_1, \cdots, v_M\}$ is a set of nodes representing the objects, and $\mathcal{E} = \{e_1, \cdots, e_K\}$ is a set of edges representing the relationships. $M$, $K$ are the number of nodes and edges.
Specifically, a node $v=(h, \mathcal{A})$ contains a head word $h$ (\textit{e.g.}, ``man'') and some attributes $\mathcal{A}=\{a_1, \cdots, a_{N_v}\}$ (\textit{e.g.}, ``white'') where $N_v$ is the number of attributes belong to the node.
An edge $e=(v_s, r, v_o)$ is a triplet which contains a subject $v_s$, a relationship $r$, and a object $v_o$ (\textit{e.g.}, (``man'', ``watching'', ``television'') ).
In practice, we adopt a scene graph parser~\cite{schuster2015generating} to transform the sentence into the language scene graph. Note that to guarantee the graph is loop-free, we slightly modified the parsing rules. According to the linguistic structure of English, \textit{i.e.}, the referent is almost the centre word modified by others, we identify the referent node as the one whose in-degree is zero.

To compute the language representation, we first represent each word $l_i$ in the given referring expression $\mathcal{L}$ by 1) embedding each word $l_i$ into an embedding vector $\bm{u}_i$, 2) applying a bi-directional LSTM to encode the embedding vectors into hidden vectors, and 3) concatenating the embedding and hidden vectors as the final representation $\bm{w}_i$, formally:
\begin{equation}
    \bm{u}_i = \mathrm{embedding}(l_i),
\end{equation}
\begin{equation}
    [\overrightarrow{\bm{h}}_i, \overleftarrow{\bm{h}}_i] = \mathrm{BiLSTM}(\bm{u}_i, [\overrightarrow{\bm{h}}_{i-1}, \overleftarrow{\bm{h}}_{i-1}]),
\end{equation}
\begin{equation}
    \bm{w}_i = \mathrm{concat}(\overrightarrow{\bm{h}}_i, \overleftarrow{\bm{h}}_i, \bm{u}_i).
\end{equation}
After that, we represent each node $v$ and edge $e$ as the meaning pooling of all containing word representations:
\begin{equation}
    \bm{w}_{v} = \mathrm{mean\_pooling}_{i \in v}(\bm{w}_i), \,\,
    \bm{w}_{e} = \mathrm{mean\_pooling}_{i \in e}(\bm{w}_i).
\end{equation}

\subsubsection{Factor Graph}
With the constructed language scene graph, it is easily to build CRF~\cite{sutton2012introduction}, thus the joint probability in Eq.~\eqref{eq:2} can be factorized as:
\begin{equation}
\label{eq:3}
    \begin{aligned}
    P(\gamma_1, \cdots, \gamma_M \mid \mathcal{G}, \mathcal{I})
    & = \prod_{v_i \in \mathcal{V}}P(\gamma_i) \prod_{(v_s, r, v_o) \in \mathcal{E}}P(\gamma_s, \gamma_o),
    \end{aligned}
\end{equation}
where $P(\gamma_i)$ denotes the grounding probability of node $v_i$, and $P(\gamma_s, \gamma_o)$  denotes the relational grounding probability of node $v_s$ and $v_o$, whose relationship is $r$.

However, it is difficult to directly model the exact probability for nodes and edges, thus we convert the CRF into a Factor Graph~\cite{kschischang2001factor}. Specifically, as shown in Figure~\ref{fig:overview}, the variables of factor graph are the nodes of the language scene graph, and the factors are composed of two types, \textit{i.e.}, the unary factors pointing to the variables, and the binary factors connecting two variables. With the factor graph, we re-write the above Eq.~\eqref{eq:3} in terms of potential functions:
\begin{equation}
\label{eq:4}
    P(\gamma_1, \cdots, \gamma_M \mid \mathcal{G}, \mathcal{I}) \propto \prod_{v_i \in \mathcal{V}}\psi_v(\gamma_i) \prod_{(v_s, r, v_o) \in \mathcal{E}}\psi_e(\gamma_s, r, \gamma_o),
\end{equation}
where $\psi_v(\gamma_i)$ is the unary potential function of node $v_i$, $\psi_e(\gamma_s, r, \gamma_o)$ is the binary potential function of the edge $(v_s, r, v_o)$.

In a nutshell, scene graph based CRF offers an inductive bias for factorizing the joint probability in Eq.~\eqref{eq:2} into the much simpler unary and binary factors in Eq.~\eqref{eq:4}. Next, we elaborate on how to compute the potential functions.

\noindent \textit{Unary Potential}.
It models how well the appearance of each region $b$ agrees with the node $v$. We use the similarity score between regions and nodes as the unary potentials:
\begin{equation}
\label{eq:5}
    \psi_v\left(
    \gamma = (v, b_i) \right) = \softmax\left(\fc(\textrm{L2norm}(\fc(\bm{x_i}) \odot \bm{w}_v))\right),
\end{equation}
where $\bm{x_i}$ is the visual feature of region $b_i$, $\bm{w}_v$ is the language representation of node $v$.
$\odot$ denotes element-wise multiplication and $\textrm{L2norm}(\cdot)$ denotes L2 normalization. Note that to maintain the non-negativity of the potentials, we use a softmax function over all regions.

\noindent \textit{Binary Potential}.
Similarly, it models the agreement between the combination $(v_s, v_o)$ of two nodes involved in the edge relation $r$:
\begin{equation}
\label{eq:6}
    \psi_e(\gamma_s= (v_s,b_i), r, \gamma_o = (v_o, b_j))
    = \softmax\left(\textrm{fc}(\textrm{L2norm}(\textrm{fc}([\bm{x}_i; \bm{x}_j]) \odot \bm{w}_e))\right).
\end{equation}
It is worth noting that our design of the binary potential preserves the directed property of scene graph, thanks to the directional feature concatenation $[\bm{x}_i; \bm{x}_j]$ in Eq.~\eqref{eq:6}.

With the scene graph based factor graphs, we can model the joint probability in Eq.~\eqref{eq:2} over all objects include referent and contexts by Eq.~\eqref{eq:4}.

\subsection{\modelspace Training \& Inference}\label{sec:3.3}
When all the grounding variables have ground-truth annotations, training the parameters of potential functions is straightforward: optimizing the unary and binary potentials of Eq.~\eqref{eq:4}. However, in visual grounding, there are no annotations for each context node and also no annotations for the edges. Therefore, we propose a marginalization strategy to make the \modelspace end-to-end trainable.

\subsubsection{Marginalization}
We propose to marginalize out all the unlabeled variables:
\begin{equation}
\label{eq:mar}
    P(\gamma_1 \mid \mathcal{G}, \mathcal{I}) = \sum_{\gamma_2}\cdots\sum_{\gamma_M}P(\gamma_1, \cdots, \gamma_M \mid \mathcal{G}, \mathcal{I}).
\end{equation}
Since directly marginalizing Eq.~\eqref{eq:mar} requires expensive computation by enumerating from $\gamma_2$ to $\gamma_M$, we adopt the standard sum-product belief propagation algorithm~\cite{kschischang2001factor} to compute the marginal probability for every node including the referent.

In brief, the algorithm works by passing messages within the graph, a variable or factor send messages to its neighbors only when it has received messages from all its other neighbors. In the beginning, we initialize the messages:
\begin{equation}
    \bm{m}^{(0)}_{i \rightarrow \alpha}(v_i) = \bm{1}, \,\,\,
    \bm{m}^{(0)}_{\alpha \rightarrow i}(v_i) = \bm{1},
\end{equation}
where $i$ and $\alpha$ are the index of variables and factors, which means $\bm{m}_{i \rightarrow \alpha}(v_i)$ denotes the messages from variables to factors, and $\bm{m}_{\alpha \rightarrow i}(v_i)$ denotes the messages from factors to variables.

After that, we pass messages by the following updating functions:
\begin{equation}
\label{eq:message_2}
    \bm{m}^{(t+1)}_{i \rightarrow \alpha}(v_i) = \prod_{j \in \mathcal{N}(i)\backslash \alpha} \bm{m}^{(t)}_{j \rightarrow i}(v_i),
\end{equation}
\begin{equation}
    \bm{m}^{(t+1)}_{\alpha \rightarrow i}(v_i) = \sum_{\alpha} \bm{\psi}(\alpha) \prod_{j \in \mathcal{N}(\alpha)\backslash i} \bm{m}^{(t)}_{j \rightarrow \alpha}(v_i),
\end{equation}
where $\mathcal{N}$ denotes the neighbors, \verb|\| denotes ``except for'', $\bm{\psi}(\alpha)$ will be $\bm{\psi}_v$ or $\bm{\psi}_v$ according to the factor types.

After $T$ times updating, we compute the beliefs for each variable, and the resultant marginals are:
\begin{equation}
    \bm{b}_i(v_i) = \prod_{\alpha \in \mathcal{N}(i)} \bm{m}^{(T)}_{\alpha \rightarrow i}(v_i),
\end{equation}
\begin{equation}
    P(\gamma_i \mid \mathcal{G}, \mathcal{I}) = \softmax_{b_i\in\mathcal{B}}(\bm{b}_i(v_i)),
\end{equation}
where we apply a softmax function to normalize the marginal probabilities. Thanks to the loop-free factor graph we constructed, when $T=2$ can we get the \textit{exact} marginals for each variable.

\subsubsection{Inference}
The marginal probability of each node is the final grounding result. Actually, the above message passing rules perform like visual reasoning for accumulating the supporting evidence for the referent. As illustrated in Figure~\ref{fig:overview},  the initial potential for ``man'' can not distinguish which man is the one ``in the red jacket''. After the message passing, the node collects the evidence from its neighbor ``jacket'' and is able to tell which region is the ``man in the red jacket''. Similarly, ``man'' also provides the supporting evidence to ``skis''. By accumulating the evidence from the message passing paths, we obtain more accurate joint grounding results for both the referent and its contexts.

\subsubsection{Training}
The training objective of our graphical model is to minimize the difference between our computed marginalized probability of the referent $P(\gamma_1 \mid \mathcal{G}, \mathcal{I})$ and the ground-truth labels.
Specifically, we use Kullback-Leibler Divergence (KLD) as the optimization function, formally:
\begin{equation}
\label{eq:8}
    \min \mathrm{KLD}\left(P(\gamma_1 \mid \mathcal{G}, \mathcal{I}), P^{gt}(\gamma \mid \mathcal{L}, \mathcal{I})\right).
\end{equation}

According to the task settings (cf. Section~\ref{sec:4.1}), the loss functions are slightly different. For the ground-truth setting, the label is a one-hot vector where only the element corresponding to the ground-truth bounding box is set to 1, otherwise 0. Therefore the loss function becomes softmax loss, formally:
\begin{equation}
    L(\Theta) = - \log P(\gamma_1 = (v_{1}, b_{gt}) \mid \mathcal{G}, \mathcal{I} ; \Theta),
\end{equation}
where $b_{gt}$ is the ground-truth region.
For the detection setting, we employ a soft label real-value vector~\cite{yu2018rethining} as:
\begin{equation}
    \begin{aligned}
    & \bm{p}^*=[p_1^*, p_2^*, \cdots, p_N^*]\\
    & p_i^*=\softmax_i\left(\{\max\{0, \mathrm{IoU}(b_i, b_{gt})-\eta\}\right).
    \end{aligned}
\end{equation}
Note that, we apply a threshold $\eta$ and a softmax function to make sure it satisfy the probability distribution. Therefore, the loss function is:
\begin{equation}
    L(\Theta) = \frac{1}{N} \sum_{i}^N p_i^* \log \bigl(\frac{p_i^*}{P(\gamma_1 = (v_{1}, b_{i}) \mid \mathcal{G}, \mathcal{I} ; \Theta)}\bigr).
\end{equation}

\section{Experiments}

\subsection{Datasets \& Evaluation Metrics}
\label{sec:4.1}
We conducted our experiments on three widely-used datasets, RefCOCO, RefCOCO+, and RefCOCOg.
\textbf{RefCOCO}~\cite{yu2016modeling} and \textbf{RefCOCO+}~\cite{yu2016modeling} both collected expression annotations with an interactive game and are split into three parts, \textit{i.e.}, validation, testA, and testB, where testA contains the images with multiple people and testB contains the image with multiple objects. The difference between them is that the location words are banned in RefCOCO+, \textit{e.g.}, ``left'', ``behind'', while those in RefCOCO are not. The average referring expression length of them is around 3.6.
\textbf{RefCOCOg}~\cite{mao2016generation} collected the referring expressions an interactive game and is split into two parts, \textit{i.e.}, validation and test. Different from RefCOCO and RefCOCO+, the average length of RefCOCOg is longer, as 8.43.
Since our work focuses on grounding referring expression based on scene graphs, we mainly did the ablations on RefCOCOg as the scene graphs from longer sentences are more qualitative.

There are two evaluation settings for different purposes. The ground-truth setting (\textbf{gt}) provides the ground-truth candidate regions and the goal is to find the best-matching region described by the referring expression. It filters out the noise from the object detector and thus we can focus on visual reasoning. We compute the percent of correctly grounded expressions as the grounding accuracy.
The detection setting (\textbf{det}) only provides an image and a referring expression, we should detect regions first. It aims to evaluate the overall performance for practical grounding systems. For det, we count a predicted region as correct if the IoU with the ground-truth is larger than 0.5.

\subsection{Implementation Details}
\label{sec:3.4}
\noindent\textbf{Visual Representation.} To extract region features of an image, we used a Faster RCNN~\cite{ren2015faster} with ResNet-101~\cite{he2016deep} as the backbone model, which pre-trained on MS-COCO images excluding the validation and test sets of RefCOCO, Ref-COCO+, and RefCOCOg.
We also incorporated the location information as relative location offsets into the region features, \textit{i.e.}, $[\frac{x_{tl}}{W}, \frac{y_{tl}}{H}, \frac{x_{br}}{W}, \frac{y_{br}}{H}, \frac{wh}{WH}]$. The final region feature dimension was 2,053.

\noindent\textbf{Language Representation.} We built scene graphs by using the Stanford Scene Graph Parser~\cite{schuster2015generating}. Unlike the previous works which usually trim the length of the expressions for computational reasons, we kept the whole sentences for more accurate scene graph parsing. For the word embedding, we used 300-d GloVe~\cite{pennington2014glove} pre-trained word vectors as initialization. For the Bi-directional LSTM, we set the number of layers as 2, and the hidden size as 1,024.

\noindent \textbf{Training Settings.} The model is trained by Adam optimizer~\cite{kingma2014adam} up to 20 epochs. The learning rate shrunk by 0.9 every 10 epochs from 1e-3. One mini-batch includes 128 images. The threshold $\eta$ was set to 0.5 according to the evaluation metrics.

\begin{table*}[t]
\centering
\begin{center}
\resizebox{\linewidth}{!}{%
\small
\begin{tabular}{ l | c | c | c | c | c | c | c | c | c }
\toprule
& & \multicolumn{3}{c|}{RefCOCO} & \multicolumn{3}{c|}{RefCOCO+} & \multicolumn{2}{c}{RefCOCOg}\\
\cline{2-10}
& setting & val & testA & testB & val & testA & testB & val & test\\
\hline
MMI~\cite{mao2016generation} & gt & - & 63.15 & 64.21 & - & 48.73 & 42.13 & - & - \\
Listener~\cite{yu2017joint} & gt & 78.36 & 77.97 & 79.86 & 61.33 & 63.10 & 58.19 & 71.32 & 71.72 \\
CMN~\cite{hu2017modeling} & gt & - & 75.94 & 79.57 & - & 59.29 & 59.34 & - & - \\
VC~\cite{zhang2018grounding} & gt & - & 78.98 & 82.39 & - & 62.56 & 62.90 & - & - \\
MAttN~\cite{yu2018mattnet} & gt & 85.65 & 85.26 & 84.57 & 71.01 & 75.13 & 66.17 & 78.10 & 78.12 \\
NMTree~\cite{liu2019learning} & gt & 85.65 & 85.63 & 85.08 & 72.84 & 75.74 & 67.62 & 78.57 & 78.21 \\
DGA~\cite{yang2019dynamic} & gt & 86.34 & 86.64 & 84.79 & 73.56 & 78.31 & 68.15 & 80.21 & 80.26 \\
CM-Att-Erase~\cite{liu2019improving} & gt & 87.47 & 88.12 & 86.32 & 73.74 & 77.58 & 68.85 & 80.23 & 80.37 \\
\hline
\textbf{Our \model} & gt & \textbf{87.98} & \textbf{88.94} & \textbf{87.06} & \textbf{74.44} & \textbf{80.03} & \textbf{69.33} & \textbf{80.51} & \textbf{80.80} \\
\hline
\hline
MMI~\cite{mao2016generation} & det & - & 64.90 & 54.51 & - & 54.03 & 42.81 & - & - \\
Listener~\cite{yu2017joint} & det & 69.48 & 72.95 & 63.43 & 55.71 & 60.43 & 48.74 & 60.21 & 59.63 \\
CMN~\cite{hu2017modeling} & det & - & 71.03 & 65.77 & - & 54.32 & 47.76 & - & - \\
VC~\cite{zhang2018grounding} & det & - & 73.33 & 67.44 & - & 58.4 & 53.18 & - & - \\
DGA~\cite{yang2019dynamic} & det & - & 78.42 & 65.53 & - & 69.07 & 51.99 & - & 63.28 \\
MAttN~\cite{yu2018mattnet} & det & 76.40 & 80.43 & 69.28 & 64.93 & 70.26 & 56.00 & 66.67 & 67.01\\
NMTree~\cite{liu2019learning} & det & 76.41 & 81.21 & 70.09 & 66.46 & 72.02 & 57.52 & 65.87 & 66.44\\
RvG-Tree~\cite{hong2019learning} & det & 75.06 & 78.61 & 69.85 & 63.51 & 67.45 & 56.66 & 66.95 & 66.51\\
DDPN~\cite{yu2018rethining} & det & 76.8 & 80.1 & 72.4 & 64.8 & 70.5 & 54.1 &  - & - \\
CM-Att-Erase~\cite{liu2019improving} & det & 78.35 & 83.14 & 71.32 & 68.09 & 73.65 & 58.03 & 67.99 & 68.67\\
\hline
\textbf{Our \model} & det & \textbf{79.27} & \textbf{84.51} & \textbf{72.05} & \textbf{68.90} & \textbf{76.06} & \textbf{59.74} & \textbf{70.96} & \textbf{71.01} \\
\bottomrule
\end{tabular}
}
\end{center}
\caption{Comparisons with stat-of-the-art visual grounding models on the RefCOCO, RefCOCO+, and RefCOCOg, with either gt bounding boxes or detected bounding boxes. Overall, our model achieves the state-of-the-art performance on all benchmarks, especially on the most challenging RefCOCOg dataset with detected regions.}
\label{table:comp}
\end{table*}

\subsection{Comparison with State-of-The-Art}
In Table~\ref{table:comp}, we compared our \modelspace to the state-of-the-arts under different experimental settings.
We can have three main observations:
1) JVGN outperforms state-of-the-art methods across all settings and datasets, especially on the RefCOCOg (the most challenging dataset with the longest sentences) and det setting (the most challenging setting), our JVGN surpasses CM-Att-Erase~\cite{liu2019improving} 2.97\% and 2.34\% on val and test sets, respectively. It indicates the efficiency of our JVGN.
2) JVGN achieves higher accuracy improvement on det setting, which is more challenging and practical than gt setting. The reason is that the referring expressions may mention some objects beyond the ground-truth regions, thus hurting the context grounding and the overall accuracy.
3) JVGN significantly outperforms other interpretable models, \textit{e.g.}, NMTree~\cite{liu2019learning} and RvG-Tree~\cite{hong2019learning}. As we will discuss in the following, JVGN also achieves better explainability, indicating JVGN breaks the well-known accuracy-explainability tradeoffs~\cite{hu2018explainable}.
In addition, it is worth noting that our model not only results efficient grounding on referents but also derives a complete grounding including contextual objects, which most existing can not provide.

\subsection{Ablation Study}
\label{sec:4.2.2}
\begin{table*}[t]
\centering
\small
\setlength{\tabcolsep}{10pt}
\begin{tabular}{@{}c|c|c|c|c|c@{}}
\toprule
\multicolumn{2}{c|}{Marginalization} & \multicolumn{2}{c|}{RefCOCOg (gt)} & \multicolumn{2}{c}{RefCOCOg (det)} \\ \cline{1-6} 
\multicolumn{1}{p{30pt}<{\centering}|}{Training} & \multicolumn{1}{p{30pt}<{\centering}|}{Inference} & \multicolumn{1}{p{20pt}<{\centering}|}{val} & \multicolumn{1}{p{20pt}<{\centering}|}{test} & \multicolumn{1}{p{20pt}<{\centering}|}{val} & \multicolumn{1}{p{20pt}<{\centering}}{test} \\
\hline
\xmark & \xmark & 78.41 & 78.32  & 69.26 & 69.55 \\
\xmark & \cmark & 78.41 & 78.31 & 69.28 & 69.54 \\
\cmark & \xmark & 79.49 & 78.73 & 70.26 & 70.45 \\
\cmark & \cmark & \textbf{80.51} & \textbf{80.80} & \textbf{70.96} & \textbf{71.01} \\
\bottomrule
\end{tabular}%
\vspace{15pt}
\caption{Ablation study results on RefCOCOg with gt and det regions. Training and Inference denote whether the model performs marginalization strategies during training and inference. With the proposed marginalization strategy, JVGN not only performs the joint visual grounding on both referent and context objects, but also gains grounding accuracy improvement.}
\label{table:ablation}
\end{table*}

\noindent \textbf{Settings.} 
We conducted ablative studies of our JVGN framework to explore the effect of marginalization training and inference strategies: 1) No marginalization during both training and inference, we set it as the baseline model. 2) Only marginalization during inference. 3) Only marginalization during training. 4) Marginalization during training and inference, it is our full JVGN model.

\noindent \textbf{Results and analyses.}
Table~\ref{table:ablation} shows the grounding result on RefCOCOg dataset. We can have the following observations:
1) Comparing with baseline, only performing marginalization during the inference can not directly lead to improvement for visual grounding. The reason is that the message passing path is blocked by the untrained binary factors.
2) Comparing with baseline, only performing marginalization during the training improves the performance. It indicates that rather than optimizing only on the referent node, the joint visual grounding benefits the 
3) The full JVGN model further surpasses the models that only performing marginalization during training or inference. With the proposed marginalization strategy, JVGN optimizes over the whole graph and infers the joint visual grounding of both referent and contextual objects. The grounding accuracy demonstrates the effectiveness of the proposed method.

\subsection{Interpretability Analysis}
Our approach is interpretable for which we conduct two experiments. \textbf{Supporting} offers an indirect quantitative result, while \textbf{human} evaluate the qualitative understandability of our complete grounding samples. In addition, this section demonstrates the \textbf{visualization} results of these samples.

\begin{figure}[t]
\begin{minipage}[b]{0.48\linewidth}
\centering
	\includegraphics[width=\linewidth]{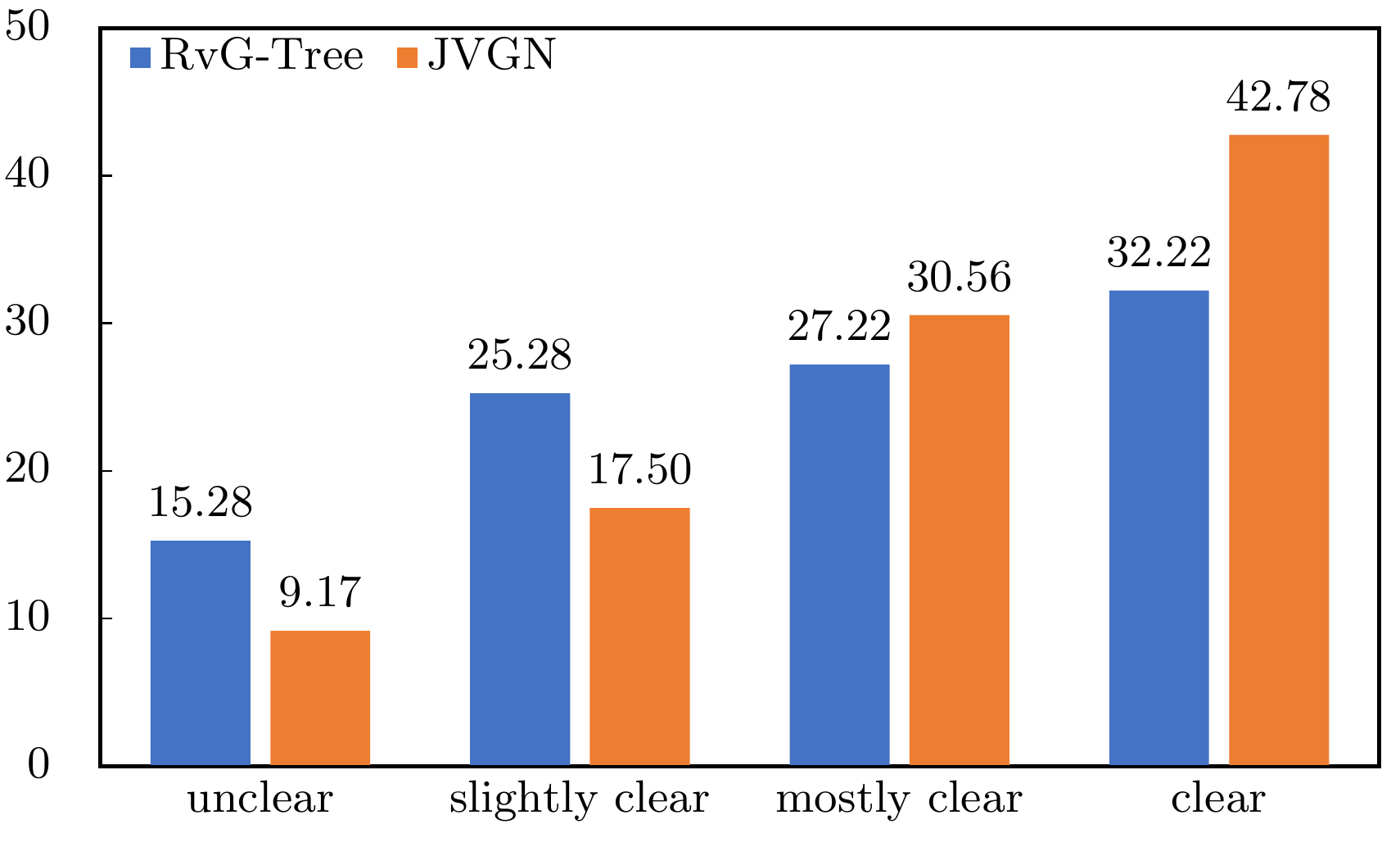}
	\caption{Human evaluations ($\%$) on JVGN and Rvg-Tree~\cite{hong2019learning}. The evaluators are asked how clearly they can understand the output results of two models. The percentage of each choice indicates the proportion of people who vote for it. Ours achieves more votes of clear.}
	\label{fig:human}
\end{minipage}
\hspace{0.04\linewidth}
\begin{minipage}[b]{0.48\linewidth}
\centering
	\includegraphics[width=\linewidth]{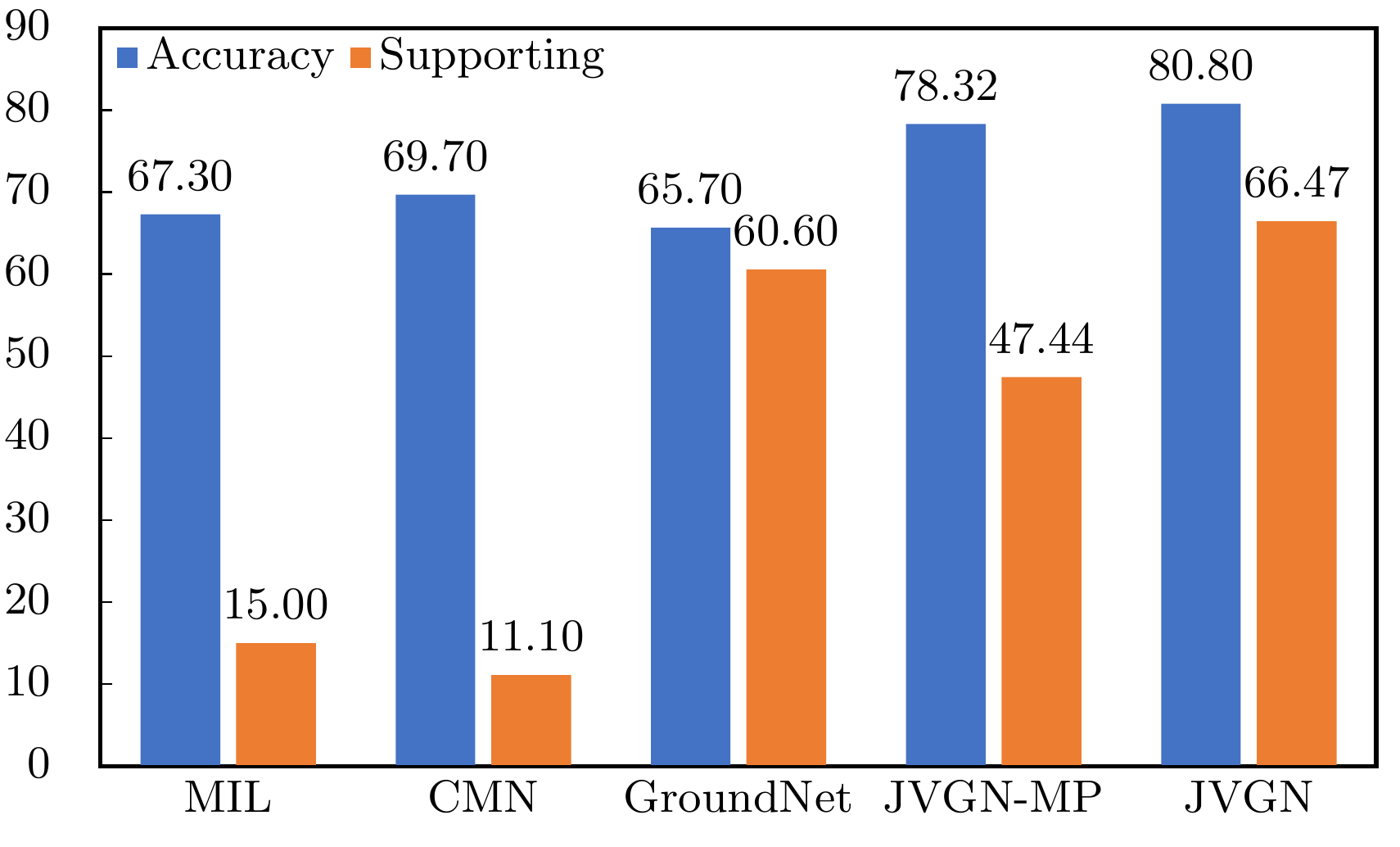}
	\caption{The accuracies ($\%$) on the referent object (in blue) and supporting objects (in orange). Message passing significantly improve the supporting accuracy. Our model achieves the highest performances compared to related works.}
	\vspace{10pt}
	\label{fig:supporting}
\end{minipage}
\end{figure}

\noindent \textbf{Supporting} denotes the object recognition experiments on the Supporting Object~\cite{cirik2018using} dataset built on RefCOCOg.
Supporting accuracy is used as an indirect evaluation of model interpretability. The intuition is that the model with the superior ability of recognizing the contextual objects (of the referent) can make more confident message passing of contexts when grounding the referent.
In Fig.~\ref{fig:supporting}, we can see that our JVGN surpasses all related methods, \textit{i.e.}, MIL~\cite{mao2016generation}, CMN~\cite{hu2017modeling}, and GroundNet~\cite{cirik2018using}, in terms of the grounding accuracies as well as the supporting accuracies. In addition, this section demonstrates the visualization results of these samples.

\begin{figure*}[t]
    \centering
    \includegraphics[width=\linewidth]{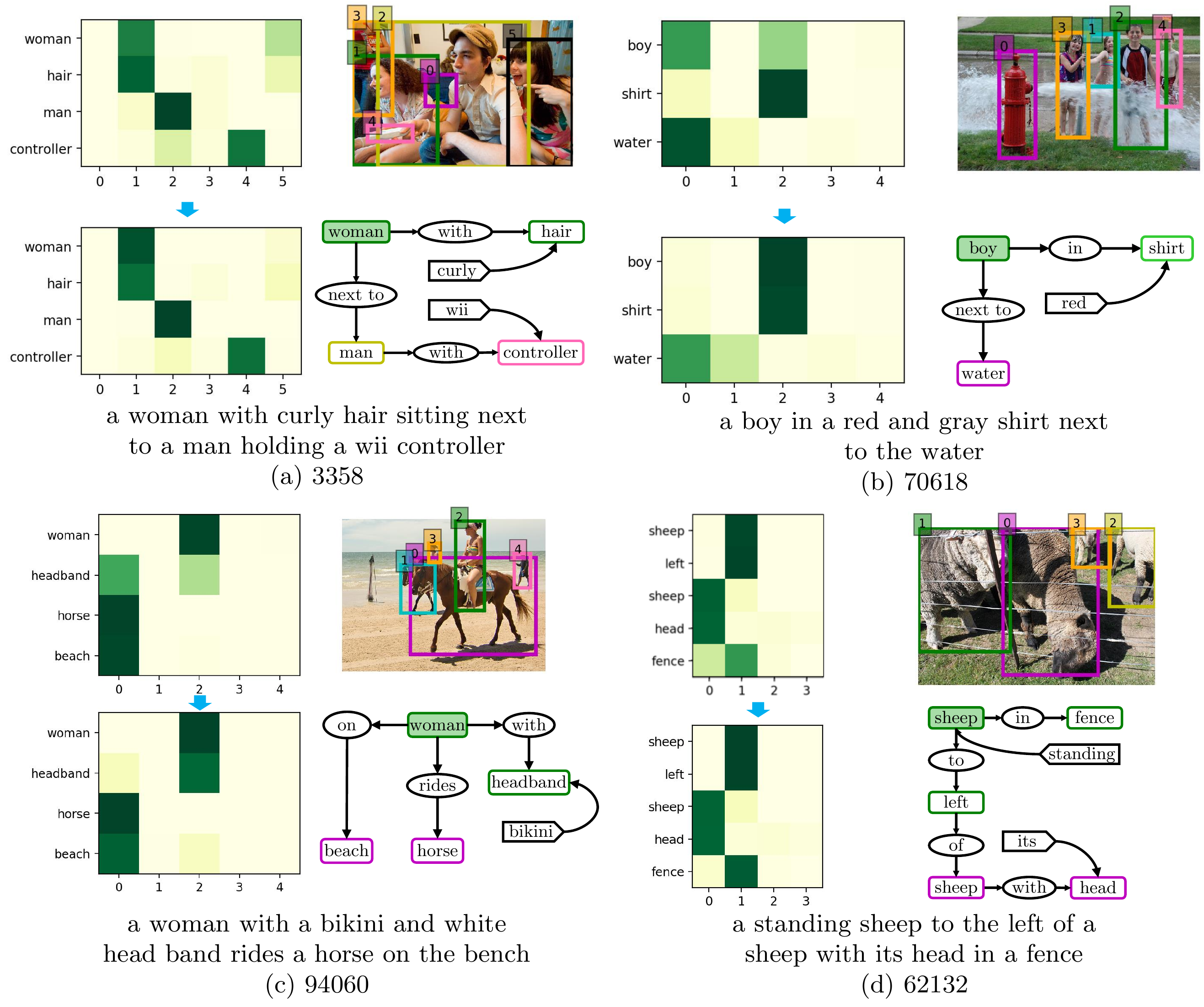}
    \caption{Visualization results on RefCOCOg. Each example includes (1) the image with regions tagged by id numbers (top right), (2) the scene graph (bottom right), (3) the initial unary potentials for every node (top left), and (4) the final marginal probability after message passing (bottom left).  As comparison, we also show one failure case (d). With the marginalization strategy, \textit{i.e.}, the message passing updating, the model performs more accurate joint visual grounding on both referent and contexts.}
    \label{fig:visualization}
\end{figure*}

\noindent \textbf{Human evaluation}
is conducted to evaluate the interpretability of \model and RvG-Tree, from a qualitative view.
We invited 12 human evaluators and each evaluator needs to rate 30 examples of each model. For each example, they are asked to judge how clearly they can understand the grounding process and rate on 4-point Likert scale, \textit{i.e.}, unclear, slightly clear, mostly clear, and clear.
For a fair evaluation, we preprocess the grounding results from different models into the same format, and shuffle the examples before showing to evaluators.
In Figure~\ref{fig:human}, we can see that our JVGN was agreed to be more understandable and interpretable than RvG-Tree by human evaluators, e.g., they rated around $73\%$ samples of our samples as clear or mostly clear, while only $59\%$ of RvG-Tree.
The more interpretability to humans makes more realistic sense to the AI systems applied in humans life. 
In the following, we show plenty of visualization results to further demonstrate the sample appearance and the inner computation of our model.

\noindent \textbf{Visualization Results}
\label{sec:4.4}
As shown in Figure~\ref{fig:visualization}, we provide some qualitative joint visual grounding results.
We can find that after updating the potentials by message passing, the likelihood becomes more concentrated (\textit{e.g.}, ``woman'' in (a)).
Even though there are some mistakes in the initial potentials, our framework can correct them after the updates (\textit{e.g.}, ``boy'' in (b)).
Not only the referent grounding becomes more accurate, but also do the context objects grounding results after message passing (\textit{e.g.}, ``headband'' in (c)).
Our \modelspace still works well on a complex graph ((\textit{e.g.}, (d)).
There are also some failure cases caused by the scene graph parsing errors (\textit{e.g.}, the object ``bikini'' is missing in (c) and ``to left of'' should be a integrated edge in (d)) or none corresponding regions (\textit{e.g.}, ``beach'' in (c) and ``fence'' in (d)).
As comparison, we also show one failure case at last.

\section{Conclusions}
In this paper, we propose a novel Joint Visual Grounding Network (\model), 
which explicitly models all the objects
mentioned in the referring expression, and hence allows the joint visual reasoning with the referent and its contexts (unlabeled in images). 
This fashion is fundamentally different from existing methods which are only able to model the holistic score between the expression and image, but lacks interpretability of the model and results.
\modelspace constructs a well-posed graphical model based on scene graphs and then marginalizes out the unlabeled contexts by message passing. On three popular visual grounding benchmarks, we showed that \modelspace is not only more high-performing on both grounding accuracy and supporting accuracy than existing methods, but also more interpretable and understandable by human evaluators. 

\renewcommand{\thefigure}{\Alph{figure}}
\renewcommand{\thetable}{\Alph{table}}
\renewcommand{\thesection}{\Alph{section}}
\setcounter{section}{0}
\setcounter{figure}{0}
\setcounter{table}{0}

\section{Appendix}
This appendix will further detail the following aspects in the main paper: 1)~Comparison with State-of-The-Art on VGG-16 (Section~\ref{sec:comp_vgg}); 2) Network Architecture (Section~\ref{sec:network}); 3) Belief Propagation Algorithm (Section~\ref{sec:algo}); and 4) More Qualitative Results (Section~\ref{sec:vis2}).

\subsection{Comparison with State-of-The-Art on VGG-16}
\label{sec:comp_vgg}
In the main paper, we have compared the proposed JVGN with recent state-of-the-art methods in Table 1. However, some earlier methods employed a VGG-16 based Faster R-CNN as the feature extractor. For a fair comparison, we further report the experimental results with the same setting in Table~\ref{table:comp_vgg}.
\begin{table*}[h]
\centering
\begin{center}
\resizebox{\linewidth}{!}{%
\small
\begin{tabular}{ l | c | c | c | c | c | c | c | c | c }
\toprule
& & \multicolumn{3}{c|}{RefCOCO} & \multicolumn{3}{c|}{RefCOCO+} & \multicolumn{2}{c}{RefCOCOg}\\
\cline{2-10}
& setting & val & testA & testB & val & testA & testB & val & test\\
\hline
MMI~\cite{mao2016generation} & gt & - & 63.15 & 64.21 & - & 48.73 & 42.13 & - & - \\
CMN~\cite{hu2017modeling} & gt & - & 75.94 & 79.57 & - & 59.29 & 59.34 & - & - \\
Speaker~\cite{yu2017joint} & gt & 78.36 & 77.97 & 79.86 & 61.33 & 63.10 & 58.19 & 71.32 & 71.72 \\
VC~\cite{zhang2018grounding} & gt & - & 78.98 & 82.39 & - & 62.56 & 62.90 & - & - \\
DGA~\cite{yang2019dynamic} & gt & 83.73 & 83.56 & 82.51 & 68.99 & 72.72 & 62.98 & 75.76 & 75.79 \\
\hline
\textbf{Our \model} & gt & \textbf{84.61} & \textbf{84.77} & \textbf{83.26} & \textbf{69.70} & \textbf{74.63} & \textbf{64.56} & \textbf{76.45} & \textbf{76.18} \\
\hline
\hline
MMI~\cite{mao2016generation} & det & - & 64.90 & 54.51 & - & 54.03 & 42.81 & - & - \\
CMN~\cite{hu2017modeling} & det & - & 71.03 & 65.77 & - & 54.32 & 47.76 & - & - \\
Speaker~\cite{yu2017joint} & det & 69.48 & 72.95 & 63.43 & 55.71 & 60.43 & 48.74 & 60.21 & 59.63 \\
VC~\cite{zhang2018grounding} & det & - & 73.33 & 67.44 & - & 58.4 & 53.18 & - & - \\
DGA~\cite{yang2019dynamic} & det & - & \textbf{78.42} & 65.53 & - & \textbf{69.07} & 51.99 & - & 63.28 \\
\hline
\textbf{Our \model} & det & \textbf{72.47} & 75.14 & \textbf{68.26} & \textbf{59.55} & 62.70 & \textbf{54.37} & \textbf{63.79} & \textbf{63.68} \\
\bottomrule
\end{tabular}
}
\end{center}
\caption{Comparisons with state-of-the-art visual grounding models on VGG-16. Note that since CMRIN~\cite{yang2019cross} used a Faster RCNN trained on Visual Genome, for the sake of fairness, we didn't compare with it. Overall, our model achieves state-of-the-art performance on most benchmarks.}
\label{table:comp_vgg}
\end{table*}

\newpage
\subsection{Network Architecture}
\label{sec:network}
In this section, we introduce the detailed network architectures of our 1) feature representations (Table~\ref{tab:network1}), 2) unary potentials (Table~\ref{tab:network2}), and 3) binary potentials (Table~\ref{tab:network3}).
\begin{table*}[!h]
\begin{center}
\begin{tabular}{|c|c|c|c|c|c|}
		\hline
		   \textbf{Index}&\textbf{Input}&\textbf{Operation}&\textbf{Output}&\textbf{Trainable Parameters}\\ \hline
		   (1)  &    -    & \,\,\,\,\,\,\,\,\,\,\, ROI feature \,\,\,\,\,\,\,\,\,\,\, & \,\,\,\, $\bm{x}_v (2048)$ \,\,\,\, & - \\ \hline
		   (2)  &    -    &  spatial feature  &  $\bm{x}_s (5)$ & - \\ \hline
		   (3)  &    -    &  one-hot word  & $l (6894)^{*}$ & - \\ \hline
		   (4)  &    (2)    &  $\mathrm{fc}(\bm{x_s})$    & $\tilde{\bm{x}}_s (512)$ & $5 \times 512 + 512$ \\ \hline
		   (5)  &    (1)(4)    &  concat $[\bm{x}_v, \tilde{\bm{x}}_s]$ & $\bm{x} (2560)$ & - \\ \hline
		   (6)  &    (3)    &  $\mathrm{embedding}(l)$ &  $\bm{u} (300)$ & $6894 \times 300$ \\ \hline
		   (7)  &    (6)    & $\mathrm{BiLSTM}(\bm{u})$ &  $\bm{h} (2048)$ & 27.64M \\ \hline
		   (8)  &    (6)(7)    & concat $[\bm{h}, \bm{u}]$ &  $\bm{w} (2348)$ & - \\ \hline
\end{tabular}
\end{center}
\caption{The details of the feature representations. $^*$ is the vocabulary size, here we take RefCOCOg as an example. For RefCOCO and RefCOCO+, the vocabulary sizes are 1991 and 2626, respectively.}
\label{tab:network1}
\end{table*}

\vspace{-20pt}

\begin{table*}[!h]
\begin{center}
\begin{tabular}{|c|c|c|c|c|c|}
		\hline
		   \textbf{Index}&\textbf{Input}&\textbf{Operation}&\textbf{Output}&\textbf{Trainable Parameters}\\ \hline
		   (1)  &    -    & \,\,\,\,\,\,\,\,\,\, visual feature  \,\,\,\,\,\,\,\,\,\, & \,\,\,\, $\bm{x} (2560)$ \,\,\,\, & - \\ \hline
		   (2)  &    -    &  node feature  & $\bm{w} (2348)$ & - \\ \hline
		   (3)  &    (1)    &  $\mathrm{fc}(\bm{x})$ & $\tilde{\bm{x}} (2348)$ & $2560 \times 2348 + 2348$ \\ \hline
		   (4)  &    (2)(3)    & multiplication($\tilde{\bm{x}}$, $\bm{w}$)  &  $\bm{m} (2348)$ & - \\ \hline
		   (5)  &    (4)    &  L2norm($\bm{m}$)  &  $\tilde{\bm{m}} (2348)$ & - \\ \hline
		   (6)  &    (5)    &  fc($\tilde{\bm{m}}$)  &  $\psi (1)$ & $2348 \times 1 + 1$ \\ \hline
\end{tabular}
\end{center}
\caption{The details of unary potential initialization function Eq.(9). Note that we omit the softmax layer over all objects.}
\label{tab:network2}
\end{table*}

\vspace{-20pt}

\begin{table*}[!h]
\begin{center}
\begin{tabular}{|c|c|c|c|c|c|}
		\hline
		   \textbf{Index}&\textbf{Input}&\textbf{Operation}&\textbf{Output}&\textbf{Trainable Parameters}\\ \hline
		   (1)  &    -    & \,\,\,\,\,\,\, subject feature  \,\,\,\,\,\,\, & \,\,\,\, $\bm{x_s} (2560)$ \,\,\,\, & - \\ \hline
		   (2)  &    -    & object feature & $\bm{x_o} (2560)$ & - \\ \hline
		   (3)  &    -    &  node feature  & $\bm{w} (2348)$ & - \\ \hline
		   (4)  &    -    &  concat $[\bm{x_s}, \bm{x_o}]$ & $\bm{x} (5120)$ & - \\ \hline
		   (5)  &    (4)    &  $\mathrm{fc}(\bm{x})$ & $\tilde{\bm{x}} (2348)$ & $5120 \times 2348 + 2348$ \\ \hline
		   (6)  &    (3)(5)    & multiplication($\tilde{\bm{x}}$, $\bm{w}$)  &  $\bm{m} (2348)$ & - \\ \hline
		   (7)  &    (6)    &  L2norm($\bm{m}$)  &  $\tilde{\bm{m}} (2348)$ & - \\ \hline
		   (8)  &    (7)    &  fc($\tilde{\bm{m}}$)  &  $\psi (1)$ & $2348 \times 1 + 1$ \\ \hline
\end{tabular}
\end{center}
\caption{The details of unary potential initialization function Eq.(10). Note that we omit the softmax layer over all relationships.}
\label{tab:network3}
\end{table*}

\newpage
\subsection{Belief Propagation Algorithm}
\label{sec:algo}
In this section, we detailed the belief propagation algorithm as follows:

\begin{algorithm}[H]
Initialize the messages
\begin{equation*}
    \bm{m}^{(0)}_{i \rightarrow \alpha}(v_i) = \bm{1}, \,\,\,
    \bm{m}^{(0)}_{\alpha \rightarrow i}(v_i) = \bm{1}.
\end{equation*}

Choose the referent node as root

Send messages from the leaves to the root and then from the root to the leaves
\begin{equation*}
    \bm{m}^{(t+1)}_{i \rightarrow \alpha}(v_i) = \prod_{j \in \mathcal{N}(i)\backslash \alpha} \bm{m}^{(t)}_{j \rightarrow i}(v_i), \,\,\,
    \bm{m}^{(t+1)}_{\alpha \rightarrow i}(v_i) = \sum_{\alpha} \bm{\psi}(\alpha) \prod_{j \in \mathcal{N}(\alpha)\backslash i} \bm{m}^{(t)}_{j \rightarrow \alpha}(v_i).
\end{equation*}

Compute the beliefs
\begin{equation*}
    \bm{b}_i(v_i) = \prod_{\alpha \in \mathcal{N}(i)} \bm{m}^T_{\alpha \rightarrow i}(v_i).
\end{equation*}

Normalize beliefs and return the marginals
\begin{equation*}
    P(\gamma_i) = \softmax(\bm{b}_i(v_i)).
\end{equation*}

\caption{Sum-Product Belief Propagation}
\end{algorithm}

\subsection{More Qualitative Results}
\label{sec:vis2}
\begin{figure*}[!ht]
    \centering
    \includegraphics[width=\linewidth]{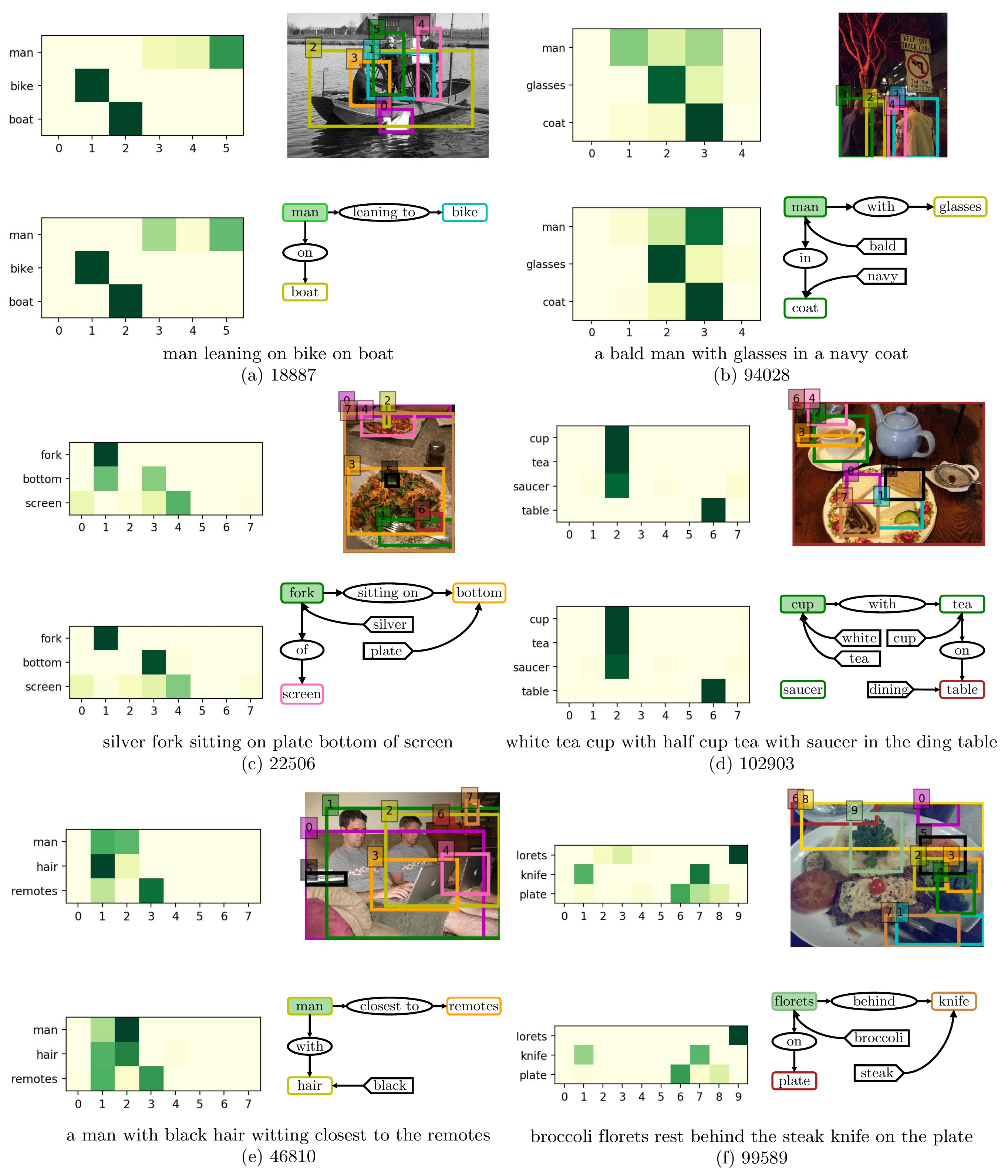}
    \caption{
    Qualitative grounding results on RefCOCOg test set. Scene graph legends: green shaded rectangle: referent node, colored rectangle: object node, arrow rectangle: attribute, oval: edge relationship. The same color of the bounding box and the node denotes a grounding. For each sample, it contains: 1) the image with regions tagged by id numbers (top right), 2) the scene graph (bottom right), 3) the initial unary potentials for every node (top left), and 4) the final marginal probability after message passing (bottom left). As a comparison, we also show failure cases (e) and (f). The sentence ID is provided for reproduction purposes. With the marginalization strategy, \textit{i.e.}, the message passing updating, the model performs more accurate joint visual grounding on both referent and contexts.}
    \label{fig:vis2}
\end{figure*}

In Fig.~\ref{fig:vis2}, we provide more qualitative results to demonstrate how the belief propagation changes potentials. As comparison, we also show two failure cases in the last row.
%
%
\bibliographystyle{splncs04}
\bibliography{citations}
\end{document}